\documentclass[a4paper]{article}

\usepackage{amsmath,graphicx}
\usepackage{xcolor}
\usepackage{color, soul, multirow}
\usepackage{amssymb}
\usepackage{subfig}
\usepackage{bm}
\usepackage{verbatim}
\usepackage{cite}
\usepackage{amssymb}
\usepackage{pifont}
\usepackage{tabularx}
\usepackage{lipsum}
\usepackage{float}
\usepackage{booktabs}
\usepackage{float}
\usepackage{tikz}
\usepackage{tikz-qtree}
\usepackage{INTERSPEECH2022}

\title{Evaluating User Perception of Speech Recognition System Quality with Semantic Distance Metric}
\vspace{0.5cm}
\name{Suyoun Kim, Duc Le, Weiyi Zheng, Tarun Singh, Abhinav Arora, Xiaoyu Zhai, \\ Christian Fuegen, Ozlem Kalinli, Michael L. Seltzer}

\vspace{0.5cm}
\address{
  Meta AI, USA}
\email{suyounkim@fb.com \\\vphantom}

\vspace{0.5cm}
\begin{document}

\maketitle
\vspace{0.5cm}
\begin{abstract}
    Measuring automatic speech recognition (ASR) system quality is critical for creating user-satisfying voice-driven applications. Word Error Rate (WER) has been traditionally used to evaluate ASR system quality; however, it sometimes correlates poorly with user perception/judgement of transcription quality. This is because WER weighs every word equally and does not consider semantic correctness which has a higher impact on user perception. In this work, we propose evaluating ASR output hypotheses quality with SemDist that can measure semantic correctness by using the distance between the semantic vectors of the reference and hypothesis extracted from a pre-trained language model. Our experimental results of 71K and 36K user annotated ASR output quality show that SemDist achieves higher correlation with user perception than WER. We also show that SemDist has higher correlation with downstream Natural Language Understanding (NLU) tasks than WER.
    
\end{abstract}

\vspace{0.5cm}
\section{Introduction}
\label{sec:intro}
\begin{figure}[t]
\vspace{0.2cm}
      \centering
      \includegraphics[width=1\columnwidth]{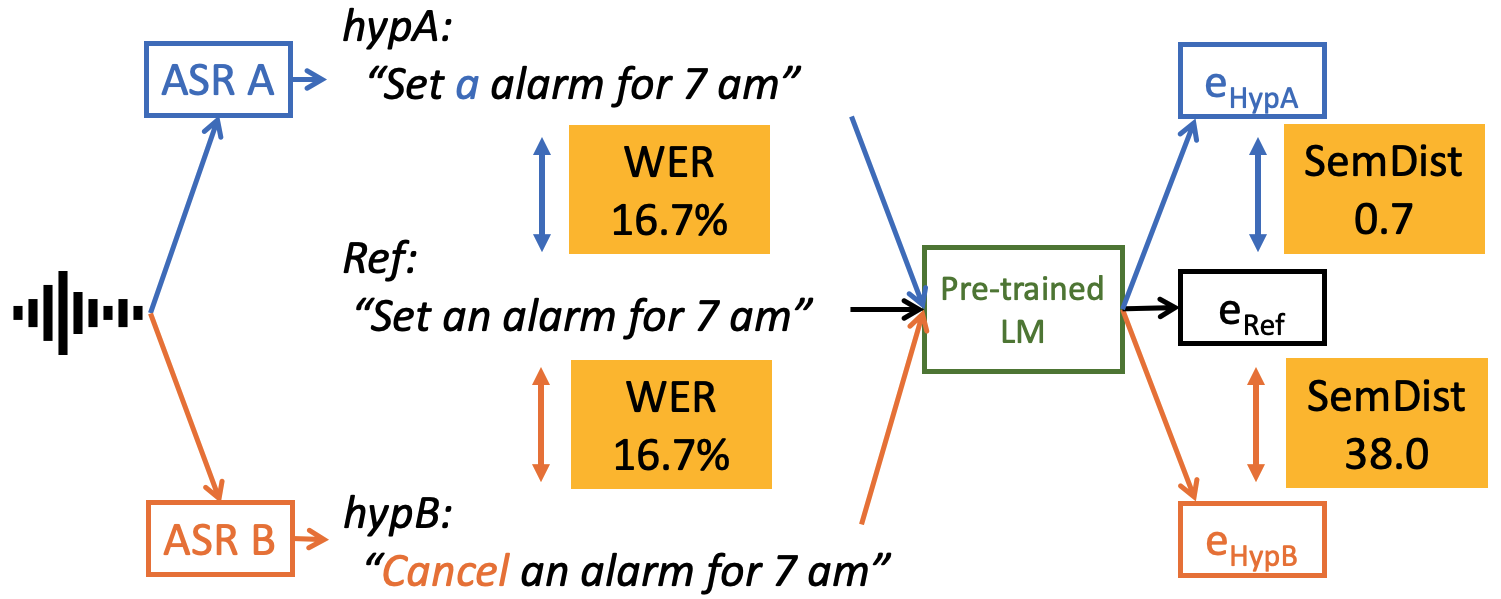}
      \caption{Comparison of two metrics, WER and SemDist, for evaluating two ASR hypotheses from different ASR systems A and B. The reference transcription is "set an alarm for 7 am". SemDist can indicate that hypothesis A is better than hypothesis B, but WER cannot.}
      \label{fig:arch}
    \end{figure}

    As voice-driven interfaces to devices become mainstream, measuring speech recognition system that can reflect user perception/judgement becomes increasingly important. Word Error Rate (WER) has been traditionally used to measure automatic speech recognition (ASR) system quality, However, it is sometimes not correlated to user perception of ASR transcription quality. This is because WER weights every word equally, and does not consider semantic correctness which has more impact on user perception. Figure \ref{fig:arch} shows an example where WER does not reflect the user perception. When the reference is ``\emph{set an alarm for 7 am}'' and two ASR hypotheses are: ``\emph{set {\color{red}a} alarm for 7 am}'' and ``\emph{{\color{red}cancel} an alarm for 7 am}'', then the former hypothesis would be preferred by the users or downstream tasks. However, WER by itself cannot identify which hypothesis is better as the error rates are identical. 
    
    Over the years, prior work has attempted to address some of WER's issues by taking word importance weight into account \cite{garofolo19991998} or adopting information retrieval to measure the performance  \cite{makhoul1999performance, hunt1990figures, mccowan2004use}. All of these metrics have been based on literal-level surface-form word correctness and are not able to measure semantic level correctness. 
    
    Meanwhile, many prior studies on transformer\cite{vaswani2017attention} based pre-trained neural language models, such as a Bidirectional Encoder Representations from Transformers (BERT), a Robustly Optimized BERT (RoBERTa), and a cross-lingual language models (XLM) \cite{Peters:2018, devlin2018bert, liu2019roberta, lample2019cross, du2020general} showed promising results in Natural Language Processing (NLP) and Natural Language Understanding (NLU) tasks. These general-purpose language models are pre-trained on billions of words, and have shown the ability to represent textual semantic information in the form of low-dimensional continuous vectors (i.e., embeddings) in textual similarity, question answering, paraphrasing, and sentiment analysis tasks \cite{devlin2018bert, liu2019roberta}. 
    
    Recently, \cite{reimers2019sentence, zhang2019bertscore, zhao2019moverscore, rei2020comet, sellam2020bleurt, yuan2021bartscore} have attempted to use these embedding features generated from the pre-trained language models to evaluate NLP/NLU systems such as machine translation and image captioning systems, and have shown their metric correlates better with human judgments and provides stronger model selection performance than existing metrics.
    Thus far, the research in measuring semantic correctness has been more focused on NLP/NLU systems. More recently, Semantic Distance (SemDist) \cite{kim2021semantic} was proposed to measure semantic correctness for ASR systems by using semantic embeddings and showed higher correlation with the downstream NLU task of intent recognition and semantic parsing, compared to WER. To the best of our knowledge, there have been no studies on user perception/judgement of ASR quality with SemDist metric.

    In this work, we first focus on studying the user perception/judgement of ASR quality using SemDist metric. We evaluated 71K and 36K user annotated ASR output quality and show that SemDist achieves higher correlation with user perception than WER.  
    Secondly, we explore a variety of strategies to compute SemDist as well. We show that the latest XLM-based \cite{lample2019cross, du2020general} token pairwise method \cite{zhang2019bertscore} performed more robustly than RoBERTa-based mean-pooling method that was used in previous work \cite{kim2021semantic}. Additionally, we show SemDist results on NLU tasks and its higher correlation than WER. Finally, we build a user perception/judgement model and show SemDist helps to estimate user perception accurately and provides insight into model selection. 

\vspace{0.5cm}
\section{Semantic Distance (SemDist)}
\label{sec:method}
    
    In this work, we measure ASR output hypothesis quality using SemDist \cite{kim2021semantic} approach with transformer-based \cite{vaswani2017attention} pre-trained LM \cite{liu2019roberta, lample2019cross}. SemDist is calculated in two steps. First, we forward the reference transcription and corresponding ASR hypothesis to the pre-trained LM and obtain semantic embeddings of the reference ($e_\text{ref}$) and hypothesis ($e_\text{hyp}$). Second, we calculate the distance between these two embeddings, ($e_\text{ref}$) and ($e_\text{hyp}$), by using cosine distance function. Although the raw value of SemDist theoretically has the same range as cosine similarity (-1 to 1), we observe that SemDist has a more limited range in practice. Thus, once we obtain SemDist, we optionally multiply it with a scalar ($\alpha = 1,000$) just for improving readability. Depending on the users’ readability preference, $\alpha$ can be chosen, and it will not affect the experimental correlation results or the model comparison results.
    
    SemDist can be obtained in various ways depending on (1) which pre-trained language model we use and (2) how to extract the embeddings. In our experiments, we compare four different SemDist: SemDist(RoBERTa-mean-pooling), SemDist(XLM-mean-pooling), SemDist(XLM-\texttt{[CLS]}), and SemDist(XLM-pairwise-token).

\vspace{0.5cm}    
    \subsection{SemDist(RoBERTa-mean-pooling)} 
    In this method, we obtain SemDist by using RoBERTa model \cite{liu2019roberta} as described in \cite{kim2021semantic}. We first obtain the semantic embeddings of the reference $e^\text{mean}_\text{ref}$ and hypothesis $e^\text{mean}_\text{hyp}$ by performing mean-pooling over all output token embeddings from RoBERTa model. We then calculate the cosine distance between $e^\text{mean}_\text{ref}$ and $e^\text{mean}_\text{hyp}$ to obtain \texttt{SemDist}.
        \begin{align}
            \label{e1}
            \texttt{SemDist} &= 1 - \text{cos\_sim}(e^\text{mean}_\text{ref}, e^\text{mean}_\text{hyp} ) \\
            \label{e2}
            &= 1 - \frac{(e^\text{mean}_\text{ref})^T\cdot e^\text{mean}_\text{hyp}}{|| e^\text{mean}_\text{ref} || \cdot || e^\text{mean}_\text{hyp} ||  }
        \end{align}
        
\vspace{0.5cm}
    \subsection{SemDist(XLM-mean-pooling)}
    The process of this method is same as SemDist(RoBERTa-mean-pooling), but we use the XLM-R model\cite{lample2019cross, du2020general} instead of RoBERTa model \cite{liu2019roberta} to extract the semantic embeddings. We can compare SemDist(XLM-mean-pooling) and SemDist(RoBERTa-mean-pooling) to see how the pre-trained LMs affect the correlation results (will be shown in \ref{sec:user_perception}).
     
\vspace{0.5cm}
    \subsection{SemDist(XLM-\texttt{[CLS]})}
    In this method, we directly use the embedding from the \texttt{CLS} token, instead of mean-pooling over all output token embeddings. The \texttt{CLS} token is a special token for the BERT-based models that is prepended to the first word and trained to hold information about the entire sentence. Once we obtain $e^\text{CLS}_\text{ref}$, $e^\text{CLS}_\text{hyp}$, we compute the cosine distance between them and obtain the SemDist(XLM-\texttt{[CLS]}). 
        \begin{align}
            \texttt{SemDist} &= 1- \text{cos\_sim}(e^\text{CLS}_\text{ref}, e^\text{CLS}_\text{hyp} )
        \end{align}
        
\vspace{0.5cm}
    \subsection{SemDist(XLM-pairwise-token)} 
    This method inspired by \cite{zhang2019bertscore}. Instead of using cosine distance between $e_\text{ref}$ and $e_\text{hyp}$ which represent the sentence-level semantic information, we use the token-level semantic information ($e^\text{tok}_\text{ref}$ and $e^\text{tok}_\text{hyp}$). We use the concept of the ``F1'' measure which is the combination of ``Precision''($p_\text{hyp}$) and ``Recall''($r_\text{ref}$) of each token of the reference. We first compute the pairwise cosine similarity between token embeddings that generated from the XLM-R model\cite{lample2019cross, du2020general} on the reference and the hypothesis. The ``Precision'' ($p_\text{hyp}$) represents the fraction of the reference token with the highest similarity score among the hypothesis tokens (in Equation \ref{e4}) and ``Recall'' ($r_\text{ref}$) is the fraction of the hypothesis token with the highest similarity score among the reference tokens (in Equation \ref{e5}). Finally, we use 1 - F1-score (the harmonic mean between $p_\text{hyp}$ and $r_\text{ref}$), $d_\text{ref}$ and $d_\text{hyp}$ as our SemDist (in Equation \ref{e3}).
        \begin{align} 
            \label{e4}
            p_\text{hyp} &= \frac{1}{|e_\text{hyp}|} \sum_{\text{tok} \in \text{hyp}} \max_{\text{tok} \in \text{ref}} (\text{cos\_sim}(e^\text{tok}_\text{ref}, e^\text{tok}_\text{hyp})) \\
            \label{e5}
            r_\text{ref} &= \frac{1}{|e_\text{ref}|} \sum_{\text{tok} \in \text{ref}} \max_{\text{tok} \in \text{hyp}} (\text{cos\_sim}(e^\text{tok}_\text{ref}, e^\text{tok}_\text{hyp})) 
        \end{align}
        \begin{align}
            \label{e3}
            \texttt{SemDist} &= 1 - F_{1} = 1 - 2 \frac{r_\text{ref} \cdot p_\text{hyp}}{r_\text{ref} + p_\text{hyp}}
        \end{align}
    
\vspace{0.5cm}
    
    Figure~\ref{fig:arch} illustrates the overall procedure to obtain SemDist with simple examples of two hypotheses A (\textit{set a alarm for 7 am}) and B (\textit{cancel an alarm for 7 am}) given the reference (\textit{set an alarm for 7 am}) and shows how SemDist and WER differ from each other. Naturally, the users or downstream tasks prefer hypothesis A over B, because A has only minor syntactic error (\emph{an} $\rightarrow$ \emph{a}) which does not hurt its meaning. As seen in this example, WER cannot separate these two hypotheses (16.7\% vs. 16.7\%) because it only measures literal word-level correctness and both hypotheses A and B has one incorrect word. However, SemDist can indicate that hypothesis A is better than B (0.7 vs. 38.0) by measuring semantic correctness.

\vspace{0.5cm}
\section{User Judgement of ASR Quality}
\label{sec:user_perception}
    We investigated the correlation of SemDist to two User Judgement tasks: HypRating and HypChoice . The detailed information of each task will be described in Section \ref{sec:data1} (HypRating) and Section \ref{sec:data2} (HypChoice).
    
    The hypotheses used in this investigation are generated from our strong baseline ASR system which is an end-to-end sequence transducer (RNN-T) \cite{Graves12transduction} with approximately 83M total parameters. The acoustic encoder is a 20-layer streamable low-latency Emformer model \cite{Shi2021emformer}. The predictor consists of three Long Short Term Memory (LSTM) layers with 512-dim hidden size, followed by 1024-dim FC projection. The joiner network contains one Rectified Linear Unit (ReLU) and one FC layer. 
    
    The ASR system is trained on 50K hours of manually transcribed data and 1.7M hours of English Facebook video data, using alignment restricted RNN-T loss \cite{Mahadeokar2021AR-RNNT} and trie-based deep biasing \cite{le21_interspeech}. Our in-house annotated evaluation set has 36k manually transcribed utterances collected from crowd-sourced workers or volunteer participants who have agreed to have their voice activity reviewed and analyzed. The evaluation set has two main domains: 21k open-domain dictation and 15k assistant-domain voice commands. 
    
    \begin{table*}[h]
        \caption{Correlation between user judgement and downstream NLU tasks and various ASR metric: CER, WER, and four different SemDist: SemDist-RM(RoBERTa-mean-pooling), SemDist-XM(XLM-mean-pooling), SemDist-XC(XLM-\texttt{[CLS]}), and SemDist-XT(XLM-pairwise-token). The Pearson correlation coefficients are reported.}
        \label{tab:corr}
        \centering
        
        \resizebox{1\textwidth}{!}{
        \centering
            \begin{tabular}{lrrrrrrrr}
            \toprule
            \textbf{Task} & \textbf{\# utter} & \textbf{Word Len.} & \textbf{CER} & \textbf{WER} & \textbf{SemDist-RM} & \textbf{SemDist-XM} & \textbf{SemDist-XC} & \textbf{SemDist-XT} \\
            \midrule
            \multicolumn{8}{l}{\textbf{User Judgement task}} \\
            UserChoice              & 36k      & 9.8       & 0.13 & 0.28 & 0.31 & 0.35 & 0.32  & \textbf{0.39} \\
            UserRating              & 71k      & 9.8       & 0.14 & 0.36 & 0.47 & 0.52 & 0.51 & \textbf{0.59} \\
            \midrule
            \multicolumn{8}{l}{\textbf{NLU task}} \\
            IntentAcc               & 10k      & 2.4       & 0.32 & 0.33    & 0.32    & \textbf{0.38}    & 0.37    & 0.37    \\
            SemanticParsing(EM)     & 10k      & 2.4       & 0.24 & 0.26    & 0.28    & 0.31    & 0.30    & \textbf{0.31}    \\
            SemanticParsing(EMTree) & 10k      & 2.4       & 0.28 & 0.29    & 0.29    & \textbf{0.34}    & 0.34    & 0.33 \\ 
            \bottomrule
            \end{tabular}
        }
    \end{table*}

    \begin{figure}[h]
\vspace{0.5cm}
      \centering
      \includegraphics[width=1\columnwidth]{ 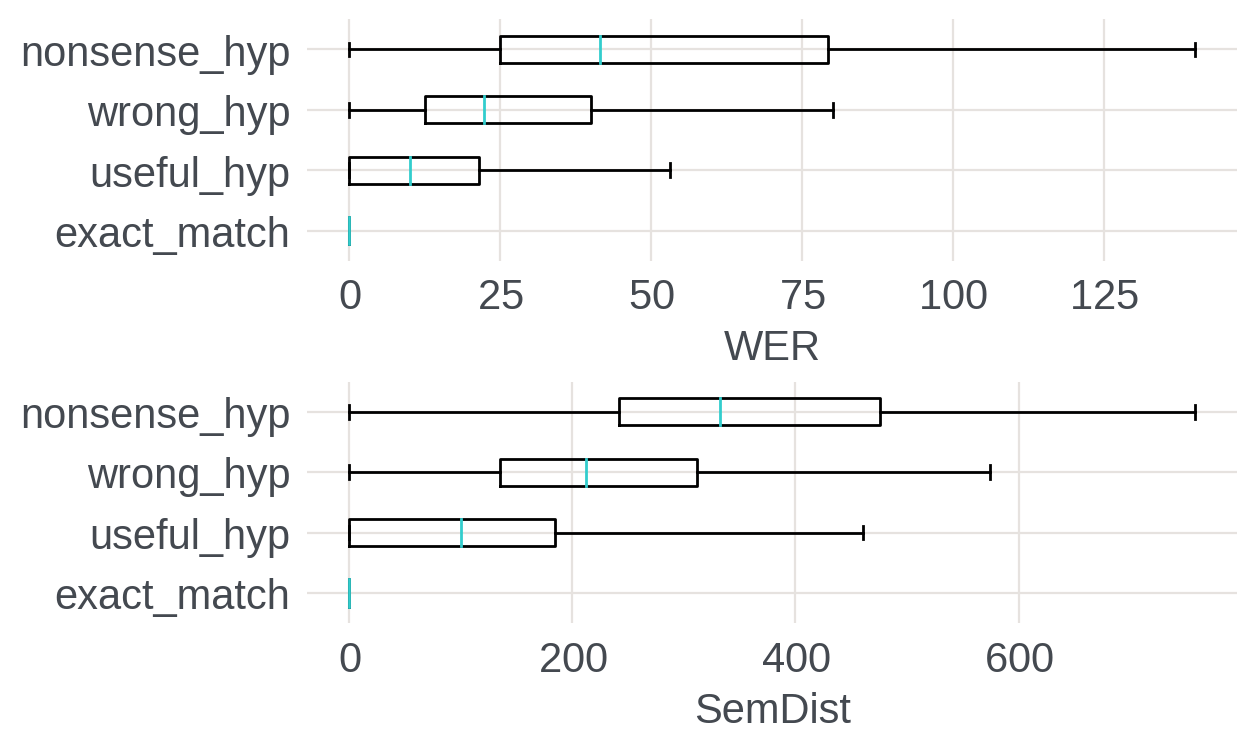}
      \caption{Comparison of the distribution of SemDist and WER for each user-rating. The box extends from the lower to upper quartile values, with a line at the median.}
      \label{fig:dist}
    \end{figure}
    
    \subsection{Hypothesis Rating (HypRating)}
    \label{sec:data1}
    HypRating consists of 73k user-ratings of ASR hypotheses. We collected the HypRating user annotation twice on our 36k evaluation set. The annotators are asked to listen to the audio and rate the hypotheses. There are four rating levels: `\texttt{exact match}', `\texttt{useful hyp}', `\texttt{wrong hyp}', `\texttt{nonsense hyp}'. A `\texttt{useful hyp}' can be thought of as a hyp which has errors, but the downstream task can still be successful. 
    In order to quantify hypothesis ratings, we assign the integer 0, 1, 2, and 3, to `\texttt{exact match}', `\texttt{useful hyp}', `\texttt{wrong hyp}', and `\texttt{nonsense hyp}', respectively. 
    
    \subsection{Side-by-Side Hypothesis Choice (HypChoice)}
    \label{sec:data2}
    HypChoice consists of 38k user annotations for ASR hypothesis pairs. The annotators are asked to listen to the audio and choose which hypothesis is better between two hypotheses A and B, and answer one of three: `\texttt{hyp A}'; A is better than B, `\texttt{hyp B}'; B is better than A, and `\texttt{equal}'; both are equally good (or bad). 
    In order to quantify the user's choice, we assign the integer -1, 1, and 0, to `\texttt{hyp A}', `\texttt{hyp B}', and `\texttt{equal}', respectively. 

\vspace{0.5cm}
\section{Downstream NLU tasks}  
\label{sec:nlu} 
    \subsection{Intent Recognition and Semantic Parsing}
    \label{sec:nlu_data}
    We next investigated the correlation of SemDist to three NLU tasks: intent recognition (IntentAcc), semantic parsing (EM), and semantic parsing (EMTree) \cite{GuptaSMKL18}. We used 10k Assistant domain ASR hypotheses that generated from our strong baseline ASR system then evaluated these hypotheses with our NLU system. The detail of the ASR and NLU system is in \cite{kim2021semantic}. Note that the size of evaluation set for this NLU task (10k) is smaller than the original assistant domain evaluation set (15k) because we selected the utterances that their annotations (i.e. intention, slot) are available. For the intent recognition task, we used 351 intent types. For the semantic parsing tasks, we used the decoupled semantic representation form \cite{AghajanyanMSDHL20} that allows the nested intents and slots. The EM is the strictest metric, which is 1 only when all the intents and the slots in the utterance are predicted correctly. The EM Tree is similar to EM but it only allows ASR errors in recognizing slot tokens.

\vspace{0.5cm}
\section{Experiments and Results}
\label{sec:exp}
    
    \begin{table*}[t]
    \centering
    \caption{Examples of Ref/Hyp that has the biggest gap between WER and SemDist. 
    }
    \label{tab:example}
    \resizebox{1\textwidth}{!}{
        \begin{tabular}{rrrl}
        \toprule
        \textbf{Gap} &\textbf{WER} & \textbf{SemDist}    & \textbf{Ref/Hyp} \\
        \midrule
        12011 & 16.67 & 0.00 & \textbf{Ref:} hey portal play \textbf{mister} blue sky\\
        & & & \textbf{Hyp:} hey portal play \textbf{\textcolor{red}{mr}} blue sky.\\
        \midrule 
        8742 & 50.00 & 0.00 & \textbf{Ref:} I smell \textbf{hot dogs}\\
        & & & \textbf{Hyp:} I smell \textbf{\textcolor{red}{hotdogs.}}\\
        \midrule 
        6390 & 6.67 & 0.01 & \textbf{Ref:} keep away it is a nightmare thank God we are separated about four thousand \textbf{kilometres}\\
        & & & \textbf{Hyp:} keep away, it is a nightmare. thank god we are separated about four thousand \textbf{\textcolor{red}{kilometers.}}\\
        \midrule 
        3807 & 20.00 & 0.02 & \textbf{Ref:} keep \textbf{time zones} in mind for the next zoom call\\
        & & & \textbf{Hyp:} keep \textbf{\textcolor{red}{timezones}} in mind for the next Zoom call.\\
        \midrule 
        2725 & 66.67 & 3.14 & \textbf{Ref:} \textbf{I'm} eagerly waiting\\
        & & & \textbf{Hyp:} \textbf{\textcolor{red}{I am}} eagerly waiting.\\
        \bottomrule
        \multicolumn{4}{c}{
        \vphantom}\\
        \multicolumn{4}{c}{(a) top-5 ($\text{Rank}_\text{WER}$ - $\text{Rank}_\text{SemDist}$) gap}\\
        & & &\\
        \multicolumn{4}{c}{
        \vphantom}\\
        \toprule
        \textbf{Gap} &\textbf{WER} & \textbf{SemDist}    & \textbf{Ref/Hyp} \\
        \midrule
        2486 & 6.25 & 302.73 & \textbf{Ref:} of course in the first time \textbf{but} one by one I able to handle those complaints \\
        & & & \textbf{Hyp:} of course, in the first time, \textbf{\textcolor{red}{birth}} one by one, I able to handle those complaints.\\
        \midrule 
        2388 & 10.00 & 473.47 & \textbf{Ref:} okay then it's kind of a great \textbf{weekly} for him\\
        & & & \textbf{Hyp:} okay, then it's kind of a great \textbf{\textcolor{red}{victory}} for him.\\
        \midrule 
        2345 & 10.00 & 426.67 & \textbf{Ref:} do you know the most \textbf{whom} he used to admire\\
        & & & \textbf{Hyp:} do you know the most \textbf{\textcolor{red}{home}} he used to admire?\\
        \midrule 
        2314 & 7.69 & 286.03 & \textbf{Ref:} sure but uh I think you will have to be patient because \textbf{usually it's} something like once a year or a little bit more but not much\\
        & & & \textbf{Hyp:} sure but uh I think you will have to be patient because \textbf{\textcolor{red}{you read}} something like once a year or a little bit more but not much.\\
        \midrule 
        2300 & 10.00 & 375.39 & \textbf{Ref:} yeah I think I've seen that in the \textbf{news} before\\
        & & & \textbf{Hyp:} yeah, I think I've seen that in the \textbf{\textcolor{red}{morning}} before\\
        \bottomrule
        \multicolumn{4}{c}{
        \vphantom}\\
        \multicolumn{4}{c}{
        (b) top-5 ($\text{Rank}_\text{SemDist}$ - $\text{Rank}_\text{WER}$) gap}\\
\vspace{0.1cm}
        \end{tabular}
    }
    \end{table*}

    \subsection{Correlation Results}
    \label{sec:res_corr}
    We first evaluated the correlation of SemDist to the various user judgement tasks and downstream NLU tasks. For calculating correlation for UserChoice, we used the subtraction of SemDist between two hyp A and B ($\text{SemDist}_\text{HypA}$ - $\text{SemDist}_\text{HypB}$), the subtraction of WER ($\text{WER}_\text{HypA}$ - $\text{WER}_\text{HypB}$), and the subtraction of CER ($\text{CER}_\text{HypA}$ - $\text{CER}_\text{HypB}$). For NLU tasks, we used 1 - IntentAcc, 1 - EM, and 1 - EMTree to consistently generate positive correlations across all tasks. 
    Table \ref{tab:corr} shows the Pearson correlation coefficients results on each tasks: User Judgement (UserChoice, UserRating) and downstream NLU (IntentAcc, SemanticParsing(EM), SemanticParsing(EMTree)) with various ASR metric: CER, WER, and four different SemDist: SemDist-RM(RoBERTa-mean-pooling), SemDist-XM(XLM-mean-pooling), SemDist-XC(XLM-\texttt{[CLS]}), and SemDist-XT(XLM-pairwise-token).
    As seen in Table \ref{tab:corr}, we observed that SemDist is significantly higher correlated to all user judgement tasks as well as downstream NLU tasks than WER. We also observed that XLM-based SemDist better correlates than RoBERTa-based SemDist with all tasks, and pairwise-token based SemDist shows the highest correlation especially with user judgement task. These results indicate that SemDist can be a better indicator for user judgement and downstream tasks than WER, and using a good semantic embedding is important.

    \subsection{How Do We Interpret SemDist Value?}
    \label{sec:interpret}
    One possible drawback to SemDist metric is that the value of SemDist is less intuitive than WER and hard to interpret. 
    To provide a better understanding of how the SemDist values should be interpreted, we show SemDist distribution for each user-rating in Figure \ref{fig:dist}. As seen this figure, the users perceive approximately 12 percent of WER and 50 of SemDist as good enough hypothesis and over 50 percent of WER and 200 SemDist as nonsense hypothesis. 

    \subsection{How Are WER and SemDist Different?}
    \label{sec:example}
    
    We next investigated how WER and SemDist evaluate differently for the same hypothesis. To do so, we first defined the gap between WER and SemDist as the change of their ranking within the entire evaluation set (36k). We assigned the ranking of WER ($\text{Rank}_\text{WER}$) and the ranking of SemDist ($\text{Rank}_\text{SemDist}$) for each utterance by sorting WER, and SemDist value. Table \ref{tab:example} (a) shows the top-5 ($\text{Rank}_\text{WER}$ - $\text{Rank}_\text{SemDist}$) gap and Table \ref{tab:example} (b) shows the top-5 ($\text{Rank}_\text{SemDist}$ - $\text{Rank}_\text{WER}$) gap. We observed that SemDist more robustly measures ASR errors by not penalizing errors that do not hurt sentence meanings (i.e. contractions and compound words) as seen in Table \ref{tab:example} (a). We also found that SemDist detects semantically nonsense word errors which occur only once within a sentence as seen in Table \ref{tab:example} (b).

    \subsection{Modeling User Judgement}
    \label{sec:predict}
    One promising aspect of SemDist is the potential to create models that can predict the user satisfaction of voice-driven applications. This can be done by training a model on pairs of SemDist and user-rating. We created three linear regression models from 71k of pairs of user-rating and (1) WER only, (2) SemDist only, and (3) both SemDist and WER. Table \ref{tab:pred} shows the comparison of $R^2$, MAE, and MSE of three models. The results show that \texttt{SemDist only} can achieve 0.35 of $R^2$ and significantly outperforms than \texttt{WER only}. Thus, using SemDist can be a promising method for estimating user satisfaction without requiring the data annotation cost. 
    
    \begin{table}[h]
        \caption{Comparison of user judgement models with three input cases: (1) WER only, (2) SemDist only, and (3) WER $+$ SemDist.}
        \label{tab:pred}
        \centering
        \resizebox{1\columnwidth}{!}{
            \begin{tabular}{rrrr}
            \toprule
                      & \textbf{WER only} & \textbf{SemDist only} & \textbf{WER $+$ SemDist} \\
            \midrule
            $R^2$ & 0.12     & 0.35         & 0.36               \\
            MAE       & 0.38     & 0.29         & 0.29               \\
            MSE       & 0.30     & 0.23         & 0.23              \\
            \bottomrule
            \end{tabular}
        }
    \end{table}

\section{Conclusion}
    \label{sec:conclusion}
    We evaluated 71k and 32k of user annotated ASR quality and showed that SemDist correlates significantly higher with user judgement than traditional metric, WER or CER. Key aspect of SemDist is measuring semantic correctness of ASR output by using semantic embeddings from the pre-trained language model for general purpose. In addition, we explored various strategies to compute SemDist and found that pairwise-token-based SemDist performs best in user judgement of ASR quality. We also showed SemDist correlates higher with downstream NLU task as well. Moreover, we demonstrated the potential of SemDist for providing insight into model selection by estimating user judgement more accurately. Moving forward, we will explore ways to use SemDist for training ASR systems.

\bibliographystyle{IEEEtran}

\bibliography{mybib}


\end{document}